\documentclass[final,5p,times,twocolumn,authoryear]{elsarticle}

\usepackage{amssymb}
\usepackage{lineno}
\usepackage[bookmarks=true,colorlinks,linkcolor=blue,urlcolor=blue,citecolor=blue]{hyperref}
\usepackage{multirow}
\usepackage{booktabs}
\usepackage{hhline}
\usepackage{colortbl}

\usepackage[ruled,vlined]{algorithm2e}
\usepackage{xcolor}
\usepackage{colortbl, booktabs}
\usepackage{arydshln}

\usepackage{times}
\usepackage{latexsym}

\usepackage[T1]{fontenc}

\usepackage[utf8]{inputenc}

\usepackage{microtype}

\usepackage{inconsolata}
\usepackage{algpseudocode}
\usepackage{graphicx}
\usepackage{multirow}
\usepackage{amsmath}
\usepackage{xcolor}
\usepackage{pifont}
\usepackage{makecell}
\usepackage{placeins}
\usepackage[table]{xcolor}
\newcommand{\zqh}{\color{black}}

\newcommand{\myred}{\textcolor{black!90!black}}
\newcommand{\mygreen}{\textcolor{green!50!black}}

\journal{Neural Networks}

\begin{document}

\begin{frontmatter}



\title{Tool Retrieval Bridge: Aligning Vague Instructions with Retriever Preferences \\via Bridge Model}

\author[a,b]{Kunfeng Chen\corref{cor0}}
\author[a,b]{Luyao Zhuang\corref{cor0}}
\author[a]{Fei Liao\corref{cor1}}
\author[a,b]{Juhua Liu\corref{cor1}}
\author[c]{Jian Wang}
\author[a,b]{Bo Du}

\address[a]{Department of Gastroenterology, Renmin Hospital, Wuhan University, Wuhan, China.}
\address[b]{School of Computer Science, Wuhan University, Wuhan, China.}
\address[c]{State Grid Corporation, China.}

\cortext[cor0]{Kunfeng Chen and Luyao Zhuang contribute equally to this work.}
\cortext[cor1]{Corresponding authors at: 
\\ Fei Liao, Department of Gastroenterology, Renmin Hospital, Wuhan University, Wuhan, China; Tel: +86-13871099860; E-mail: feiliao@whu.edu.cn.
\\ Juhua Liu, School of Computer Science, Wuhan University, Wuhan, China; Tel: +86-18062452253; E-mail: liujuhua@whu.edu.cn.}

\begin{abstract}
    Tool learning has emerged as a promising paradigm for large language models (LLMs) to address real-world challenges. Due to the extensive and irregularly updated number of tools, tool retrieval for selecting the desired tool subset is essential. However, current tool retrieval methods are usually based on academic benchmarks containing overly detailed instructions (\textit{e.g.}, specific API names and parameters), while real-world instructions are more vague. Such a discrepancy would hinder the tool retrieval in real-world applications. In this paper, we first construct a new benchmark, \texttt{VGToolBench}, to simulate human vague instructions. Based on this, we conduct a series of preliminary analyses and find that vague instructions indeed damage the performance of tool retrieval. To this end, we propose a simple-yet-effective \textbf{Tool Retrieval Bridge} (\texttt{TRB}) approach to boost the performance of tool retrieval for vague instructions. The principle of \texttt{TRB} is to introduce a bridge model to rewrite the vague instructions into more specific ones and alleviate the gap between vague instructions and retriever preferences.
    \myred{We conduct extensive experiments under multiple commonly used retrieval settings, and the results show that \texttt{TRB} effectively mitigates the ambiguity of vague instructions while delivering consistent and substantial improvements across all baseline retrievers. For example, with the help of \texttt{TRB}, BM25 achieves a relative improvement of up to 111.51\%, \textit{i.e.}, increasing the average NDCG score from 9.73 to 19.59.}
    The source code and models are publicly available at https://github.com/kfchenhn/\texttt{TRB}.
\end{abstract}




\end{frontmatter}


\section{Introduction}
\label{sec_intro}

{\zqh
While large language models (LLMs)~\citep{achiam2023gpt,touvron2023llama,bai2023qwen} have achieved great success in the NLP community, they still fall short in terms of complex computations and providing the latest information, due to their reliance on fixed and parametric knowledge~\citep{qu2025tool}. To this end, tool learning~\citep{lu2023chameleon, cai2024large, hsieh2023tool, toolmeetllm, ding2025toolcoder, 2025QinTool, shi2025toollearning}, aiming to unleash the power of LLMs by integrating with external tools, has attracted enormous attention. For instance, by using search engines, LLMs can obtain more accurate and timely information, thus better interacting with the external world.

\begin{figure}[t]
    \centering
    \includegraphics[width=0.85\columnwidth]{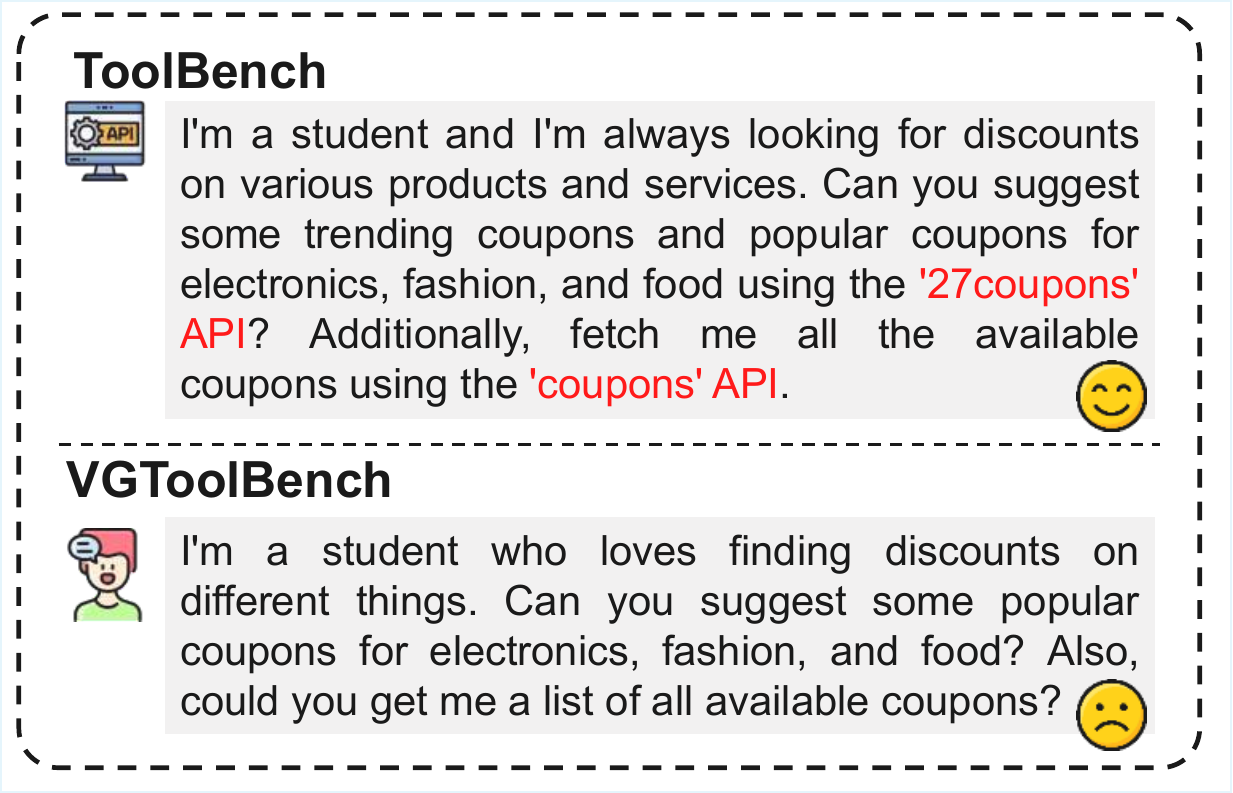}
    \caption{
    {\textbf{Instruction Comparison between ToolBench~\citep{qin2023toolllm} and our \texttt{VGToolBench}.} As seen, ToolBench contains more detailed and specific instructions (highlighted in \textcolor{red}{red}), while the instructions of our \texttt{VGToolBench} are vague and more aligned with real-world scenarios.
    }
    }
    \label{fig:begin}
\end{figure}
One of the key challenges of tool learning is to determine specific tools for corresponding tasks~\citep{xu2024enhancing}. To achieve this goal, there are two lines of research: 1) In-Context Learning (ICL)~\citep{brown2020language,lu2023chameleon,gupta2023visual,suris2023vipergpt}, which instructs the LLMs to understand the tool descriptions in the context and select the related tools~\citep{gao2024confucius}; 2) Supervised Fine-Tuning (SFT), which incorporates tool learning capabilities into model parameters through fine-tuning~\citep{schick2023toolformer,wang2024toolgen,hao2023toolkengpt}. Since the SFT of LLMs is usually computationally expensive and memory-intensive, current works generally focus on ICL-based methods. However, the vast number of tools makes it difficult for LLMs to contain all tools within the limited input length. Thus, tool retrieval, selecting appropriate tools from a large-scale tool set, is essential for tool learning.

Recently, to facilitate tool retrieval and usage, a series of tool learning benchmarks have been proposed, such as ToolBench~\citep{qin2023toolllm} and APIBench ~\citep{peng2022revisiting}. While containing large-scale and high-quality tools, these benchmarks usually differ from real-world scenarios~\citep{wu2024toolplanner}, as illustrated in Figure~\ref{fig:begin}. Specifically, most of them are automatically constructed by proprietary LLMs, \textit{e.g.}, ChatGPT, and tend to be overly detailed and specific. 
\myred{Conversely, instead of specifying the required exact tools or APIs, human users usually describe their goals in broad and outcome-focused terms. That is, users' queries mainly focus on the final results, \textit{e.g.}, ``help me organize this data'' or ``find out why my device is not working'', which are obviously different from the detailed benchmark queries. Such a style gap might reduce the robustness and effectiveness of tool-learning models in real-world applications.}

To this end, we first construct a vague tool learning benchmark, denoted as \texttt{VGToolBench}, to simulate real-world applications. Specifically, we prompt a third-party LLM to automatically rewrite the original instructions of ToolBench~\citep{qin2023toolllm} into vague ones. 
\myred{This design follows the observation that real-world user queries tend to focus on goals rather than specific tool names, so the rewritten instructions retain only high-level intents.}
Based on \texttt{VGToolBench}, we conduct a series of preliminary analyses and reveal that vague instructions indeed significantly damage the performance of tool retrieval, \textit{i.e.}, leading up to \textbf{50.39\%} relative performance drops. In response, we further propose a simple-yet-effective \textbf{Tool Retrieval Bridge} (namely \texttt{TRB}) to boost the performance of tool retrieval for vague instructions. The principle of our \texttt{TRB} is to introduce a bridge model for rewriting vague instructions, thus alleviating the gap between vague instructions and retriever preferences. More specifically, the training of the bridge model consists of two-stage processes: \ding{182} we fine-tune the model to encourage it to rewrite the vague instructions into more specific ones, and \ding{183} the SFT model is further optimized to align the retriever preference by using the retrieval performance as the reward for reinforcement learning (RL). \myred{For training the bridge model, we keep the data scale aligned with ToolBench to ensure stable and reliable training.
}

We evaluate \texttt{TRB} on the constructed \texttt{VGToolBench} among four widely-used retrievers, including BM25~\citep{robertson2009probabilistic}, TF-IDF~\citep{ramos2003using}, AdaEmbeding and ToolRetriever~\citep{qin2023toolllm}. Extensive experimental results show that \texttt{TRB} not only effectively alleviates the negative effect of vague instructions (\textit{i.e.}, bringing up to \textbf{+111.51\%} relative average performance gains against the vanilla retriever), but also brings consistent and significant improvements across all retrievers. 
Moreover, we perform comprehensive and in-depth analyses to investigate the effect of key components in \texttt{TRB}. More encouragingly, the results indicate that the iterative RL strategy can further boost the effectiveness of \texttt{TRB}, and our \texttt{TRB} is also beneficial to the subsequent tool usage.


To summarize, our contributions are three-fold: (1) We construct a new tool learning benchmark, namely \texttt{VGToolBench}, to simulate the vague instructions in real-world scenarios. (2) To alleviate the negative effect of vague instructions, we propose a simple-yet-effective Tool Retrieval Bridge (\texttt{TRB}) approach, which employs a bridge model to align the instructions with retriever preferences. (3) Extensive experiments demonstrate that \texttt{TRB} surpasses the vanilla retrievers by a large margin (\textit{i.e.}, up to \textbf{+111.51\%} relative average gains), proving its effectiveness.

The rest of this paper is organized as follows. In Section~\ref{sec:related}, we briefly review the related works. In Section~\ref{sec:datacon_preexp}, we construct \texttt{VGToolBench} and conduct preliminary experiments. Section~\ref{sec:method} introduces our proposed method in detail. Section~\ref{sec:experiments} reports and analyzes our experimental results. Lastly, we conclude our study in Section~\ref{sec:conclusion}.

}

\begin{table*}[t]
\centering
\small
\caption{The prompt template for vague instruction generation}
\setlength{\arrayrulewidth}{0.2mm} 
\setlength{\tabcolsep}{3pt} 
\begin{tabular}{|p{\linewidth}|}
\hline
Given a user instruction and a list of related APIs, your task is to generate a fuzzier version of the user instruction by simplifying or replacing technical terms with synonyms, without changing the user's core requirements. You can follow these steps:\\
- Analyze the user instruction: identify how many tasks the user has and what the specific needs are.\\
- Compare with the API list: match the user's tasks with the relevant APIs. Remove any references to specific APIs or redundant technical details.\\
- Simplify technical terms: replace highly specialized or technical terms with more common, everyday language. The goal is to make the instruction sound like something a regular user would say in casual conversation.\\
- Rephrase the instruction: use simpler language or synonyms where appropriate, but ensure the core intent remains unchanged.\\
- Output the result: provide the final fuzzier version of the instruction.\\
\\
Example1:\\
\textbf{Original instruction:} I'm organizing a gaming tournament for my company's employees. Could you provide the statistics and ratings of highly skilled players in popular games like Dota 2 and World of Tanks? Also, recommend some gaming peripherals and accessories for the event.\\
\textbf{Relevant APIs:} [tool\_name: World of Tanks Stats, api\_name: Get Stats], 
[tool\_name: DOTA 2 Steam Web, api\_name: Match History], 
[tool\_name: DOTA 2 Steam Web, api\_name: Match Details], 
[tool\_name: CheapShark - Game Deals, api\_name: List of Deals], 
[tool\_name: CheapShark - Game Deals, api\_name: Game Lookup]\\
\textbf{Answer:} I'm organizing a company gaming tournament and need player stats for top players in popular games. Can you also recommend some good gaming gear for the event?\\
\\
Example2: \\
\textbf{Original instruction:} I want to surprise my family with a personalized playlist. Can you recommend some popular tracks from different genres? Additionally, provide me with the detailed information of a playlist that I want to create. Also, fetch the track URL of a specific song that I want to include in the playlist.\\
\textbf{Relevant APIs:} [tool\_name: Shazam, api\_name: artists/get-summary], [tool\_name: Deezer, api\_name: Track], [tool\_name: Soundcloud, api\_name: /playlist/info] \\
\textbf{Answer:} I want to make a special playlist for my family. Can you suggest some hit songs from different music styles? Also, give me more info about the playlist I'm putting together. Finally, can you get me the link to a specific track I want to add?\\
\\
Now, please make the fuzzier instruction.\\ 
\textbf{Original instruction:} \texttt{\{instruction\}}\\
\textbf{Relevant APIs:} [\texttt{\{APIs\}}]\\
\textbf{Answer:}\\
\hline
\end{tabular}
\label{tab:fuzz_prompt}
\end{table*}

\section{Related Works}
\label{sec:related}

\subsection{\textbf{Tool Learning Benchmarks}}
In recent years, the community has built diverse benchmarks to promote research on tool learning, including APIBench~\citep{patil2023gorilla}, API-Bank~\citep{li2023api}, ToolAplaca~\citep{tang2023toolalpaca}, ToolQA~\citep{zhuang2023toolqa}, and ToolBench~\citep{qin2023toolllm}, \textit{etc}. \myred{It is noteworthy that constructing reliable evaluation benchmarks often shares methodological similarities with dataset design in other NLP areas, such as social-media text classification, abusive-language detection, and feature-selection-based categorization~\citep{a-omar2021multi, b-khairy2021automatic, c-omar2023quantum, d-el2024detecting, f-omar2020comparative, g-farghaly2020building, h-farghaly2020developing, j-mamdouh2023high, k-mamdouh2022new}, where data diversity and annotation quality likewise play a crucial role.} However, except for ToolBench, most benchmarks generally suffer from limited API quantity and diversity, and their application scenarios are subject to significant limitations. For example, ToolAplaca~\citep{tang2023toolalpaca} only contains 400 tools, 400 APIs, and 3,938 instructions, with each instruction involving only one tool.
In contrast, ToolBench~\citep{qin2023toolllm} provides a high-quality dataset containing 3,451 tools, including 16,464 real-world APIs collected from RapidAPI Hub. To simulate real-world scenarios, ToolBench prompts ChatGPT to generate nearly 200K qualified (instruction, relevant API) pairs, which are divided into three categories, \textit{i.e.}, single-tool instructions (\textbf{I1}), intra-category multi-tool instructions (\textbf{I2}), and intra-collection multi-tool instructions (\textbf{I3}). 
However, since the instruction generation of ToolBench primarily relies on the names and descriptions of these APIs, these instructions tend to be too detailed and include specific API information that deviates from human preference. Furthermore, these instructions often contain API information, making it easier for the tool retriever to identify the relevant tools~\citep{qian2024tell}, which may not reflect the tool retrieval performance in real-world scenarios. To this end, ToolPlanner~\citet{wu2024toolplanner} selects the intra-category multi-tool subset (\textbf{I2}) of ToolBench as the seed dataset, and generates new multi-granularity user instructions by directly trimming all tool-related information and GPT-4 prompts, thereby constructing a new benchmark MGToolBench. 

Different from MGToolBench, we build a new benchmark \texttt{VGToolBench}, which is more natural and closer to real-world scenarios. Specifically, we use the entire ToolBench as the seed dataset and prompt GPT-4o to remove tool-related information of instructions. More importantly, \texttt{VGToolBench} is built for tool retrieval, while MGToolBench is used for SFT tool learning.

\subsection{\textbf{Tool Learning Framework}}

Tool learning aims to equip LLMs with external tools to enhance and expand their capabilities~\citep{ruan2023tptu, huang2024planning, wang2024llms}. Existing frameworks can be broadly classified into two categories: In-Context Learning-based (ICL-based) frameworks~\citep{brown2020language,lu2023chameleon,gupta2023visual,gao2023pal,hsieh2023tool,suris2023vipergpt} and Supervised Fine-Tuning-based (SFT-based) frameworks~\citep{schick2023toolformer,hao2023toolkengpt, wang2024toolgen, zeng-etal-2024-agenttuning}. Specifically, ICL-based frameworks encourage LLMs to use tools with descriptions or documentation~\citep{patil2023gorilla, qin2023toolllm, zheng2024toolrerank, yuan2024easytool}, while SFT-based frameworks focus on training LLMs directly on specialized tools datasets to develop tool usage capabilities~\citep{hao2023toolkengpt, tang2023toolalpaca, gao2024confucius}. 
In practice, due to the limited input length of LLMs, it is almost impossible to input the names and descriptions of over thousands of tools into LLMs, whether it is ICL- or SFT-based framework. Therefore, it is crucial to equip LLMs with a tool retrieval component to select appropriate tools from a large-scale tool set. Recent works have proposed various methods for tool retrieval, \textit{e.g.}, using sparse retrieval techniques like BM25~\citep{patil2023gorilla,liu2024lost} or dense methods~\citep{qin2023toolllm,kong2024tptu,gao2023precise} that rely on sentence embeddings of the Sentence-BERT transformer model trained on instruction-tool document pair datasets. 

Despite their effectiveness, in our preliminary experiments in Section \ref{sec:pre_exp}, we observe that these methods still suffer from some limitations, especially when facing vague user instructions. In response to this issue, we propose a new tool retrieval pipeline \texttt{TRB} that incorporates a bridge model to boost the retrieval performance of tool retrievers for vague instructions. \myred{Notably, across broader AI applications, similar benefits are obtained from structured external knowledge or auxiliary modules~\citep{e-shams2024dynamic, l-mahmoud2013framework, m-badawy2021topic}, echoing the motivation behind equipping LLMs with core components.}

\section{Dataset Construction and Preliminary Experiments}
\label{sec:datacon_preexp}

\subsection{\textbf{Task Definition}}
The task of tool retrieval aims to select a subset of tools relevant to the user's instruction from a large-scale tool repository. Mathematically, the user's instruction is represented as $q$, and the tool repository is denoted by $\mathcal{D} = \{d_1, d_2, \dots, d_M\}$, where each $d_i$ represents the description of a specific tool, and $M$ denotes the total number of tools. For each instruction $q$, there is a ground truth list of relevant tools $U_g \subseteq \mathcal{D}$. The retriever $R$ computes the relevance score $R(q, d_i)$ between the instruction and each tool description $d_i$, and selects the top-$k$ tools with the highest relevance scores. As a result, we can obtain the final tool subset, denoted as $U^k_r \subseteq \mathcal{D}$, which is subsequently used to perform the tool usage process.


\subsection{\textbf{\texttt{VGToolBench} Construction}}
\label{sec:vague_generation}

To evaluate the performance of existing tool retrievers for vague user instructions, we adopt ToolBench as the seed dataset and construct a new tool learning benchmark (namely \texttt{VGToolBench}), which can better simulate human vague instructions in real-world scenarios. Inspired by many prior studies~\citep{wu2024toolplanner,zhong2024iterative}, we also attempt to make full use of the LLM's powerful instruction-following and in-context learning abilities to automatically convert the original instructions into vague ones. Specifically, we design a prompt template to instruct the powerful proprietary LLM, \textit{i.e.}, GPT-4o, to perform the instruction rewriting. As shown in the prompt template of Table~\ref{tab:fuzz_prompt}, given the original instructions and their associated tool metadata (\textit{i.e.}, API names, parameters, and descriptions) of ToolBench, GPT-4o is forced to remove the detailed and explicit tool information of instruction $q$, and convert it into a more vague version, denoted as $q'$. Notably, we additionally use two examples as demonstrations in the prompt to guide the data generation.

For each instruction $q$ of ToolBench, we perform the above data processes and obtain the required vague instruction $q'$. Finally, we can construct the \texttt{VGToolBench} benchmark, denoted as $\mathcal{S}$, which contains $|\mathcal{S}|$ triple training data, \textit{i.e.}, $\mathcal{S}=\{(q'_i,q_i,U_{g_i})\}^N_{i=1}$.


\myred{Following ToolBench, \texttt{VGToolBench} is organized into three subsets (\textit{i.e.}, \textbf{I1}, \textbf{I2}, and \textbf{I3}), which correspond to progressively more complex instruction–tool interaction scenarios. Specifically, \textbf{I1} comprises instructions that require invoking a single tool. \textbf{I2} includes instructions that involve multiple tools within the same functional category.  Additionallly,  \textbf{I3} consists of instructions that necessitate coordinating multiple tools across different tool collections, thereby posing a more challenging multi-tool reasoning setting. The overall dataset statistics, including the number of training and test instances for each subset, are summarized in Table~\ref{tab:dataset_stats}.}

\begin{table}[ht]
    \centering
    \caption{Statistics of \texttt{VGToolBench}.}
    \label{tab:dataset_stats}
    \resizebox{0.7\columnwidth}{!}{
    \begin{tabular}{lccc}
    \toprule
    \textbf{Dataset} & \textbf{I1} & \textbf{I2} & \textbf{I3} \\
    \midrule
    \midrule
    \textit{\#Training} & 86,826 & 84,425 & 25,151 \\
    \textit{\#Test }    & 600    & 400    & 100 \\
    \bottomrule
    \end{tabular}}
\end{table}

\subsection{\textbf{Impact of Vague Instructions on Tool Retrieval}} 
\label{sec:pre_exp}

\paragraph{\textbf{Setting}} 

Based on our proposed \texttt{VGToolBench}, we conduct some preliminary experiments to investigate the impact of vague instructions on tool retrieval. Specifically, we select four widely-used tool retrievers and conduct comparative experiments on \texttt{VGToolBench} and ToolBench. Notably, the detailed experimental settings can be seen in Section \ref{sec:setting}.

\paragraph{\textbf{Results}} 

\begin{figure}[t]
    \centering
    \includegraphics[width=\columnwidth]{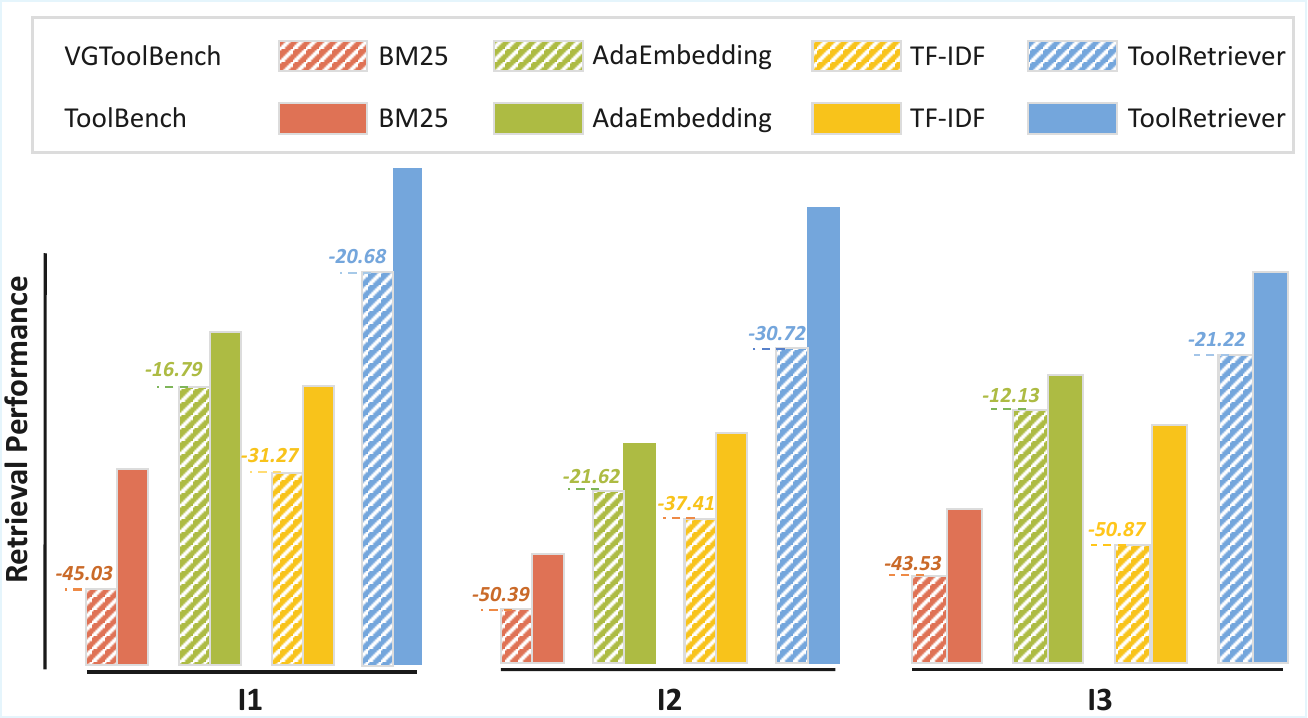}
    \caption{\textbf{Retrieval Performance comparison of \texttt{VGToolBench} \textit{v.s.} ToolBench.} The x-axis denotes the different types of sub-sets in \texttt{VGToolBench} and ToolBench, and the y-axis denotes the tool retrieval performance, evaluated by the average of NDCG@5 and NDCG@10, where the evaluation details can be found in Section~\ref{sec:setting}. The numerical results represent the relative decrease compared to the results on Toolbench. We can observe that the performance declines significantly, \textit{e.g.}, the performance drops on BM25 is up to \textbf{50.39 \%}.}
    
    \label{fig:pre_exp}
\end{figure}

The comparative results are shown in Figure~\ref{fig:pre_exp} and Table~\ref{tab:main_exp_new}. 
From these results, we empirically observe that: 

\textbf{1) It is more challenging for all retrievers to capture the semantic connection between user vague instructions and tool descriptions.} As seen, all retrievers perform worse on \texttt{VGToolBench} than TooBench. Although ToolRetriever~\citep{qin2023toolllm} fine-tuned on ToolBench performs best on the original ToolBench benchmark, it still falls short in dealing with vague instructions, \textit{e.g.}, leading up to \textbf{-30.72\%} relative performance drops on \textbf{I2} subset of \texttt{VGToolBench}. 

\textbf{2) Simpler retrievers struggle more with vague instructions.} The performance of sparse retrievers drops more significantly than dense retrievers, \textit{e.g.}, BM25 drops by \textbf{-50.39\%} on \textbf{I2} and TF-IDF drops by \textbf{-50.87\%} on \textbf{I3}. One of the possible reasons is that sparse retrievers depend more on the overlap between the instruction and tool description. Thus, in our \texttt{VGToolBench}, without the detailed tool description in instructions, the sparse retrievers often fail in obtaining the relevant tools. 

In fact, the user's instructions in real-world scenarios are often complex, ambiguous, and dynamic. It contrasts with the detailed preferences of tool retrievers, which typically perform better when provided with more specific instructions that include terms directly related to tool information. Therefore, the question arises: \textit{\textbf{How can we improve performance when handling vague instructions?}}

\section{Aligning Vague Instructions with Retriever Preferences}
\label{sec:method}

\begin{figure*}[t]
    \centering
    \includegraphics[width=0.8\textwidth]{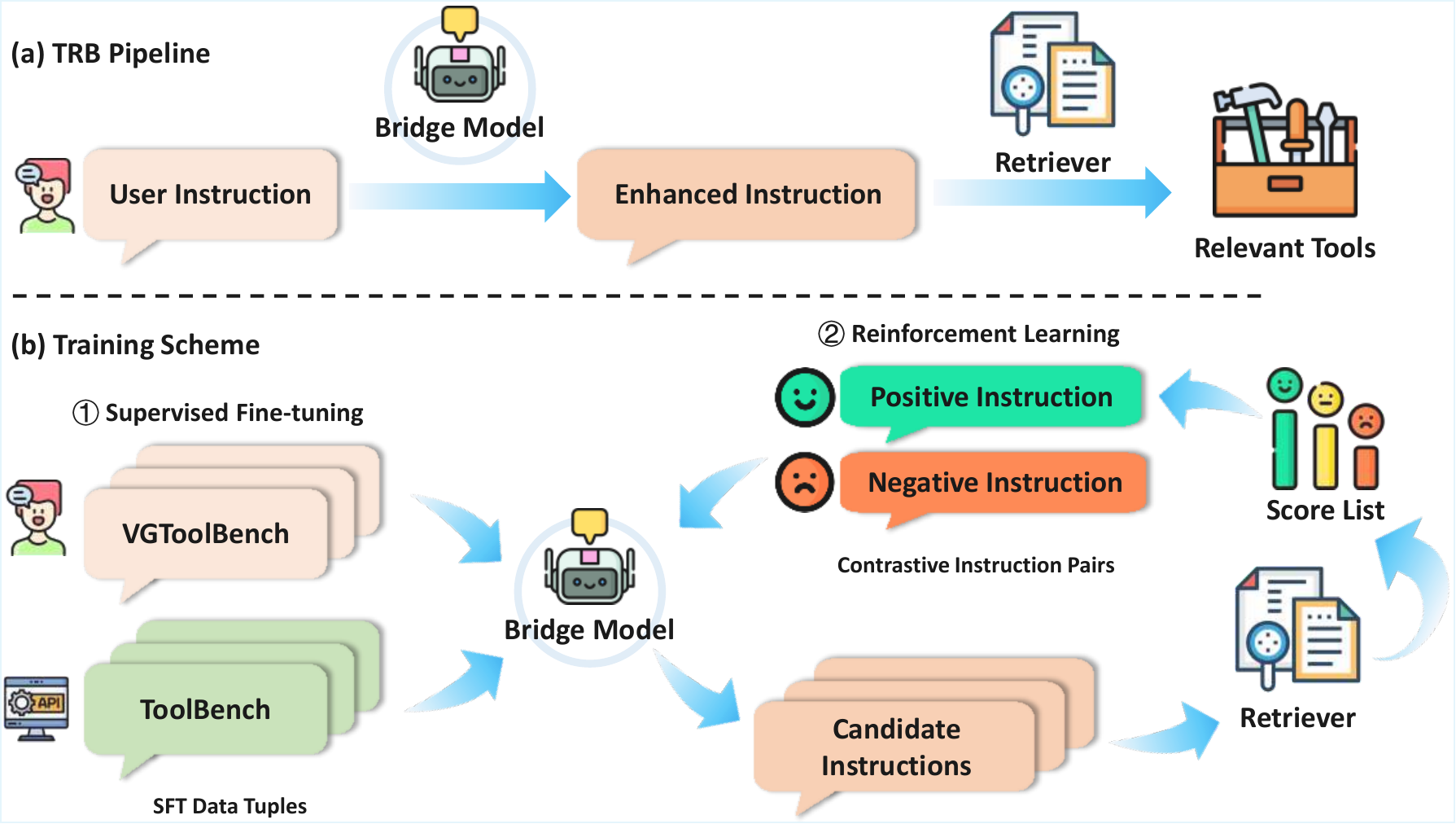}
    \caption{
    \textbf{Overview of our proposed \texttt{TRB}.} (a) The pipeline of \texttt{TRB}, where the core is to introduces a bridge model to enhance the instruction into a more specific version. (b) The training scheme of the bridge model, which consists of a two-stage process: \ding{182} \textbf{Supervised Fine-tuning}, \ding{183} \textbf{Reinforcement Learning}. 
    \myred{We first construct paired input–output instances by aligning data correspondences between \texttt{VGToolBench} and ToolBench, and train an initial model through supervised fine-tuning. Building on this initial model, we further apply reinforcement learning to align its preference by using the retrieval performance as reward signal.
    }
    } 
    \label{fig:method}
\end{figure*}

\begin{algorithm*}[t!]
\DontPrintSemicolon
\small
\KwIn{$\mathcal{S}=\{(q',q,U_g)^{(i)}\}_{i=1}^{|\mathcal{S}|}$: the warm-up SFT  dataset, $U_{g_i}$: the ground truth tools for both $q'_i$ and $q_i$, $N$: the number of data sampling, $\beta$: hyperparameter in DPO, $\pi_{\theta}$: the initial bridge model.}
\KwOut{Final bridge model $\pi_{\theta}$}
\textcolor{blue}{\texttt{// Supervised Fine-tuning}} \;

Optimize $\theta$ with warm-up SFT training objective: 
$\mathcal{L}_{\text{SFT}} = - \mathbb{E}_{(q'_i, q_i) \sim \mathcal{S}} \left[\sum^{T_i}_{t=1} \log \pi_\theta(q_{i_t} | q_{i_{<t}}, q'_i) \right]$

\textcolor{blue}{\texttt{// RL Optimization}}\;

    $\pi_\mathrm{ref}=\pi_\theta$ \;
    
    \For{ $q'_i \in \mathcal{S}$}{
    Get $N$ candidate instructions $\hat{q}_{ij} \sim \pi_\mathrm{\theta}(q'_i)$,
        
        \For{ $j\in\{1,2,\ldots,N\}$ }{
            \For{ $k\in\{5,10\}$ }{
            Retrieve the top-$k$ most relevant tools:      
            $U_{r_{ij}}^k = \texttt{Top-k}\left( \{R(\hat{q}_{ij}, d_m)\}_{m=1}^{M} \right)$,
            }
            Calculate the relevance score:
            $Score_{ij} = \frac{\text{NDCG}(U^5_{r_{ij}}, U_{g_i})+ \text{NDCG}(U^{10}_{r_{ij}}, U_{g_i})}{2}$
        }
        Sort $N$ candidate instructions based on $Score_{ij}$ and get the positive-negative pair: $\hat{q}^+_{i} \succ \hat{q}^-_{i} \mid q'_i$ \;
    }

    Construct the contrastive dataset:
    $\mathcal{S}_{dpo} = \{ (q'_i, \hat{q}^+_{i}, \hat{q}^-_{i}) \}_{i=1}^{|\mathcal{S}_{dpo}|}.$ \;
    
    Optimize $\theta$ with the DPO training objective: $\mathcal{L}_{\text{DPO}} =  - \mathbb{E}_{(q'_i, \hat{q}^+_{i}, \hat{q}^-_{i}) \sim \mathcal{S}_{dpo}} \left[ \log \sigma \left( \beta \log \frac{\pi_{\theta}(\hat{q}^+_{i} \mid q'_i)}{\pi_{\text{ref}}(\hat{q}^-_{i} \mid q'_i)} - \beta \log \frac{\pi_{\theta}(\hat{q}^+_{i} \mid q'_i)}{\pi_{\text{ref}}(\hat{q}^-_{i} \mid q'_i)} \right) \right]$ \;

\Return{$\pi_{\theta}$}
\caption{\textbf{Training scheme of bridge model in \texttt{TRB}.}}
\label{algo:main}
\end{algorithm*}

\paragraph{\textbf{Motivation}} 

Based on the above observations, we recognize that the key to improving the retrieval performance of vague instructions is to bridge the gap between user habits and retriever preferences. Motivated by this, we propose the Tool Retrieval Bridge method, denoted as \texttt{TRB}, which effectively boosts the retriever's performance via rewriting vague instructions into more detailed ones, thereby alleviating the gap. The overview of \texttt{TRB} is illustrated in Figure~\ref{fig:method}.     

\paragraph{\textbf{The Pipeline of \texttt{TRB}}} 

The enhancement process of \texttt{TRB}, illustrated in Figure \ref{fig:method} (a), takes the user instruction $q'_i \in \mathcal{S}$ as input and rewrites it into an detailed one $\hat{q_i}$ via our proposed bridge model. Then, a relevant set of tools, \textit{i.e.}, $U^k_{r_i}$, is retrieved based on the detailed instruction $\hat{q_i}$. In short, this process aims to clarify user intent and restructure the vague instructions into more specific ones. By doing so, the bridge model can alleviate the gap between the vague instructions and tool retrievers, thus resulting in better performance.

\paragraph{\textbf{Training Scheme of Bridge Model}}

As aforementioned, the core of our \texttt{TRB} is the bridge model that aims to enhance the vague instructions. In this part, we introduce the training scheme of this bridge model in details. As shown in Figure \ref{fig:method}(b), the training scheme mainly contains two techniques: \ding{182} Supervised Fine-Tuning (SFT), which aims to warm up the bridge model by using the the original instruction in ToolBench as the ground-truth; \ding{183} Reinforcement Learning (RL), which aims to further reinforce the bridge model by leveraging the feedback from the retriever. The overall training scheme is presented in Algorithm~\ref{algo:main}.

\textbf{\ding{182} Supervised Fine-Tuning.} 
First, given the training data of \texttt{VGToolBench} $\mathcal{S}=\{(q'_i,q_i,U_{g_i})\}^N_{i=1}$, we encourage the bridge model to learn how to rewrite the vague instruction $q'_i$ into a more informative one. Here, we simply use the original instruction $q_i$ as the ground-truth, and train the bridge model using a standard log-likelihood objective function:
\begin{align}
\mathcal{L}_{\text{SFT}} = - \mathbb{E}_{(q'_i, q_i) \sim \mathcal{S}} \left[\sum^{T_i}_{t=1} \log \pi_\theta(q_{i_t} | q_{i_{<t}}, q'_i) \right].
\end{align}
where $\pi_\theta$ denotes the parameters of bridge model, $T_i$ is the token length of instruction $q_i$, $\mathcal{L}_{\text{SFT}}$ is the loss function of SFT training. This warm-up SFT training allows the bridge model to learn how to follow the prompt and enhance the instructions, thus building the foundation for subsequent RL optimization.

\textbf{\ding{183} Reinforcement Learning.} 
\myred{Reinforcement learning is a training paradigm that can be effectively employed for preference optimization and has been widely adopted across a variety of applications~\citep{zhang2024adp, stojanovic2023fault}.} While SFT could effectively train the bridge model, it lacks feedback regarding retrieval performance. Depending entirely on SFT does not result in optimal performance, as proven by our empirical results in Table~\ref{tab:ablation_sft}. Thus, to better align the bridge model with retriever preferences, we further perform the RL optimization on the warm-up SFT model. Specifically, we adopt the widely-used Direct Preference Optimization (DPO)~\citep{rafailov2023direct} algorithm for continued training. Notably, we also attempt to use the other online RL algorithms, \textit{e.g.}, PPO~\citep{schulman2017proximal} and GRPO~\citep{shao2024deepseekmath}. Considering the intolerable training latency of these methods, we finally use the DPO for a better trade-off between training efficiency and performance\footnote{The analysis of different RL algorithms can be found in Section~\ref{sec:ablation}.}. The key of DPO is to construct the pairwise training data. Thus, in the following content, we first introduce the pairwise data construction process, and then present the training details.


\textbf{Pairwise Data Construction.} During the sampling phase, for each vague instruction $q'_i$, we use the SFT model $\pi_\theta$ (initialized from the warm-up model) to generate $N$ candidate enhanced instructions:
\begin{equation}
\hat{q}_{ij} \sim \pi_\theta(q'_i),
\end{equation}
where $\hat{q}_{ij}$ denotes the $j$-th enhanced instruction for $i$-th training data, and $j\in\{1,2,\ldots,N\}$. For each candidate instruction $\hat{q}_{ij}$, the retriever $R$ computes the similarity between the instruction and each tool in the tool repository, retrieving the top-$k$ most relevant tools:
\begin{equation}
U_{r_{ij}}^k = \texttt{Top-k}\left( \{R(\hat{q}_{ij}, d_m)\}_{m=1}^{M} \right).
\end{equation}

Each candidate instruction is then scored using the NDCG~\citep{jarvelin2002cumulated} metric to quantitatively measure the alignment between the predicted $U^k_{r_{ij}}$ and the ground-truth tools $U_{g_i}$, \textit{i.e.}, NDCG($U^k_{r_{ij}}$, $U_{g_i}$). The final score for each instruction is the average NDCG score computed for $k\in\{5, 10\}$:
\begin{equation}
Score_{ij} = \frac{\text{NDCG}(U^5_{r_{ij}}, U_{g_i})+ \text{NDCG}(U^{10}_{r_{ij}}, U_{g_i})}{2},
\end{equation}
where $Score_{ij}$ is the relevance score for $U^k_{r_{ij}}$ and can reflect the effectiveness of the bridge model, as a large score refers to better performance.
Next, we select the contrastive instruction pair ($\hat{q}^+_{i}, \hat{q}^-_{i}$) with the highest and lowest scores from the candidate instructions. Note that we only retain instruction pairs with differing rewards. Finally, we obtain the contrastive instruction dataset $\mathcal{S}_{dpo}$ as follows:
\begin{equation}
    \mathcal{S}_{dpo} = \{ (q'_i, \hat{q}^+_{i}, \hat{q}^-_{i}) \}_{i=1}^{|\mathcal{S}_{dpo}|}.
\end{equation}

\textbf{Training.} During this phase, we compute the contrastive loss with the contrastive instruction dataset to assist the bridge model in learning from high-score instructions:
\begin{equation}
\resizebox{\columnwidth}{!}{$
\begin{aligned}
\mathcal{L}_{\text{DPO}} =  - \mathbb{E}_{(q'_i, \hat{q}^+_{i}, \hat{q}^-_{i}) \sim \mathcal{S}_{dpo}} \left[ \log \sigma \left( \beta \log \frac{\pi_{\theta}(\hat{q}^+_{i} \mid q'_i)}{\pi_{\text{ref}}(\hat{q}^-_{i} \mid q'_i)} - \beta \log \frac{\pi_{\theta}(\hat{q}^+_{i} \mid q'_i)}{\pi_{\text{ref}}(\hat{q}^-_{i} \mid q'_i)} \right) \right],
\end{aligned}
$}
\end{equation}
where $\pi_{\text{ref}}$ is the reference model, which is initialized from the SFT model and kept fixed during the training. $\sigma$ is the \texttt{Sigmoid} function, and $\beta$ is the hyperparameter. This optimization objective aims to increase the likelihood of the instructions with better retrieval performance $\hat{q}^+_{i}$ and decrease the likelihood of instruction with poorer retrieval performance $\hat{q}^-_{i}$.


\section{Experiments}
\label{sec:experiments}

\subsection{\textbf{Setup}} 
\label{sec:setting}
\paragraph{\textbf{Datasets}}

In addition to the preliminary experimental results in Section~\ref{sec:pre_exp}, we provide a more comprehensive validation and analysis in this section. To verify the effectiveness of our \texttt{TRB}, we conduct extensive experiments on both ToolBench~\citep{qin2023toolllm} and our proposed \texttt{VGToolBench}. Specifically, ToolBench consists of 16,464 APIs spanning various categories. Based on these APIs, it provides nearly 200k high-quality (instruction, relevant API) pairs as a dataset for tool learning, which is further divided into three subsets: single-tool instructions ({\bf I1}), intra-category multi-tool instructions ({\bf I2}), and inter-category multi-tool instructions ({\bf I3}). \texttt{VGToolBench} is constructed by rewriting the instructions in ToolBench, ensuring a one-to-one correspondence between each instruction in ToolBench and its rewritten counterpart in \texttt{VGToolBench}. We use the instructions from both ToolBench and \texttt{VGToolBench} to retrieve relevant tools from the complete tool set and evaluate their retrieval performance. Notably, our \texttt{TRB} method is applied to enhance the instructions given to the retrievers. 

\paragraph{\textbf{Evaluation}} 

Following many prior studies~\citep{qin2023toolllm, zheng2024toolrerank}, we utilize the popular Normalized Discounted Cumulative Gain (\textbf{NDCG})~\citep{jarvelin2002cumulated} metric to evaluate the performance of tool retrieval.
Specifically, we report NDCG@$k$ scores for $k \in \{5, 10\}$, which measure the ranking performance of retrieved tools by considering the positions of relevant tools within the top-$k$ results. Higher NDCG@$k$ scores refer to better retrieval performance. \myred{Notably, except specified, the retrieval performance of subsequent experiments is calculated by the average of  NDCG@5 and  NDCG@10. For tool calling task, the performance is reported in terms of accuracy, where only cases with entirely correct tool calling are counted as correct.}

\begin{table}[t]
    \centering
    \caption{\textbf{Training hyperparameters used in \texttt{TRB}.}}
    \label{tab:hyperparameter}
    \resizebox{0.9\columnwidth}{!}{
    \begin{tabular}{lcc}
    \toprule
    Hyperparameter & \makecell{Stage 1(SFT)} & \makecell{Stage 2 (RL)} \\
    \midrule
    \midrule
    epoch & 3 & 3 \\
    batch\_size & 16 & 12 \\
    learning\_rate & 5e-5 & 2e-5 \\
    warmup\_ratio & 0.0 & 0.02 \\
    optimizer & Adam & Adam \\
    max\_sequence\_length & 1024 & 1024 \\
    GPUs & 1 & 2 \\
    \myred{Number of candidate samples (N)} & \myred{-} & \myred{4} \\
    \bottomrule
    \end{tabular}}
\end{table}

\begin{table*}[t]
    \centering
    \caption{\textbf{Experimental results on ToolBench and \texttt{VGToolBench}.} ``Avg.'' means the average performance of NDCG@5 and NDCG@10. "$\%\Delta\downarrow$" represents the relative performance degradation (\%) of \texttt{VGToolBench} compared to the original ToolBench, while "$\%\Delta\uparrow$" indicates the relative performance improvement (\%) of the four tool retrievers equipped with our proposed \texttt{TRB} on \texttt{VGToolBench}. \myred{\bf Red} results indicate that the retrievers' performance on \texttt{VGToolBench} degrades compared to ToolBench, while \mygreen{\bf green} results denote that the retrievers equipped with \texttt{TRB} improves its performance on \texttt{VGToolBench}. We observe that \texttt{TRB} consistently outperforms the baselines in both single-tool and multi-tool situations scenarios.}   
    \resizebox{\textwidth}{!}{  
    \begin{tabular}{ccccccccccc}
    \toprule
    \multirow{2}{*}{Methods} & \multirow{2}{*}{Datasets} & \multicolumn{3}{c}{I1} & \multicolumn{3}{c}{I2} & \multicolumn{3}{c}{I3}  \\ 
    \cmidrule(lr){3-5}  \cmidrule(lr){6-8}  \cmidrule(lr){9-11}
    & & NDCG@5 & NDCG@10 & Avg. & NDCG@5 & NDCG@10 & Avg. & NDCG@5 & NDCG@10 & Avg. \\ 
    \midrule
    \midrule
    \multirow{2}{*}{BM25} & ToolBench  &\underline{22.73} &\underline{24.55} &\underline{23.64}  &\underline{17.55}  &\underline{19.85}  &\underline{18.70}   &\bf 24.73  &\bf 27.75  &\bf 26.24   \\
    & \texttt{VGToolBench}  &  \myred{12.49}   &\myred{13.50}  &\myred{13.00}   &\myred{8.81}   &\myred{9.73}   &\myred{9.27}   &\myred{13.94}  &\myred{15.70}  &\myred{14.82}     \\
    &\%$\Delta\downarrow$   &\cellcolor{gray!20}{\myred{-45.05}}  &\cellcolor{gray!20}{\myred{-45.01}} &\cellcolor{gray!20}{\myred{-45.03}}   &\cellcolor{gray!20}{\myred{-49.80}}  &\cellcolor{gray!20}{\myred{-50.98}}  &\cellcolor{gray!20}{\myred{-50.39}}   &\cellcolor{gray!20}{\myred{-43.63}} &\cellcolor{gray!20}{\myred{-43.42}}  &\cellcolor{gray!20}{\myred{-43.53}}  \\
    \hdashline
    \textbf{\texttt{+\texttt{TRB}}}  & \texttt{VGToolBench} &\mygreen{\bf 23.32}  &\mygreen{\bf 25.29} &\mygreen{\bf 24.31}   &\mygreen{\bf 19.06}    &\mygreen{\bf 20.11}   &\mygreen{\bf 19.59}   &\mygreen{18.54}    &\mygreen{21.97}    &\mygreen{20.26}   \\
    &\%$\Delta\uparrow$    &\cellcolor{gray!20}{\mygreen{86.71}}  &\cellcolor{gray!20}{\mygreen{87.33}} &\cellcolor{gray!20}{\mygreen{87.02}}   &\cellcolor{gray!20}{\mygreen{116.35}}  &\cellcolor{gray!20}{\mygreen{106.68}}  &\cellcolor{gray!20}{\mygreen{111.51}}   &\cellcolor{gray!20}{\mygreen{33.00}} &\cellcolor{gray!20}{\mygreen{39.94}}  &\cellcolor{gray!20}{\mygreen{36.47}}   \\
    \hline

    \multirow{2}{*}{TF-IDF} & ToolBench  &\bf 46.12 &\bf 49.44  &\bf 47.78   &\bf 37.93  &\bf 40.57  &\bf 39.25  &\bf 37.54  &\bf 43.59  &\bf 40.57    \\
    & \texttt{VGToolBench}  &\myred{31.50}   &\myred{34.19}  &\myred{32.85}   &\myred{23.60}  &\myred{25.54}  &\myred{24.57}   &\myred{17.64}  &\myred{22.35}  &\myred{20.00}   \\
    &\%$\Delta\downarrow$   &\cellcolor{gray!20}{\myred{-31.70}}  &\cellcolor{gray!20}{\myred{-30.85}} &\cellcolor{gray!20}{\myred{-31.27}}   &\cellcolor{gray!20}{\myred{-37.78}}  &\cellcolor{gray!20}{\myred{-37.05}}  &\cellcolor{gray!20}{\myred{-37.41}}   &\cellcolor{gray!20}{\myred{-53.01}} &\cellcolor{gray!20}{\myred{-48.73}}  &\cellcolor{gray!20}{\myred{-50.87}}    \\
    \hdashline
    \textbf{\texttt{+\texttt{TRB}}} & \texttt{VGToolBench} &\underline{\mygreen{43.23}}   &\underline{\mygreen{46.44}}    &\underline{\mygreen{44.84}}    &\underline{\mygreen{34.56}}   &\underline{\mygreen{37.15}}    &\underline{\mygreen{35.86}}   &\underline{\mygreen{27.79}}    &\underline{\mygreen{34.38}}    &\underline{\mygreen{31.09}}   \\
    &\%$\Delta\uparrow$ &\cellcolor{gray!20}{\mygreen{37.24}}  &\cellcolor{gray!20}{\mygreen{35.83}} &\cellcolor{gray!20}{\mygreen{36.53}}   &\cellcolor{gray!20}{\mygreen{46.44}}  &\cellcolor{gray!20}{\mygreen{45.46}}  &\cellcolor{gray!20}{\mygreen{45.95}}   &\cellcolor{gray!20}{\mygreen{57.54}} &\cellcolor{gray!20}{\mygreen{53.83}}  &\cellcolor{gray!20}{\mygreen{55.68}} \\
    \hline

    \multirow{2}{*}{AdaEmbedding} & ToolBench  &\bf 54.91   &\bf 59.32  &\bf 57.12   &\bf 36.18  &\bf 38.93  &\bf 37.56   &\bf 45.78  &\bf 52.45  &\bf 49.12   \\
    & \texttt{VGToolBench}  &\myred{45.15}   &\myred{49.95}  &\myred{47.55}   &\myred{27.84}  &\myred{31.07}  &\myred{29.46}   &\myred{40.62}  &\myred{45.64}  & \myred{43.13}  \\
    &\%$\Delta\downarrow$   &\cellcolor{gray!20}{\myred{-17.77}}  &\cellcolor{gray!20}{\myred{-15.80}} &\cellcolor{gray!20}{\myred{-16.79}}   &\cellcolor{gray!20}{\myred{-23.05}}  &\cellcolor{gray!20}{\myred{-20.19}}  &\cellcolor{gray!20}{\myred{-21.62}}   &\cellcolor{gray!20}{\myred{-11.27}} &\cellcolor{gray!20}{\myred{-12.98}}  &\cellcolor{gray!20}{\myred{-12.13}}    \\
    \hdashline
    \textbf{\texttt{+\texttt{TRB}}} & \texttt{VGToolBench} &\underline{\mygreen{52.27}}   &\underline{\mygreen{56.53}}    &\underline{\mygreen{54.40}}   &\underline{\mygreen{32.53}}    &\underline{\mygreen{35.90}}    &\underline{\mygreen{34.22}}   &\underline{\mygreen{45.41}}    &\underline{\mygreen{50.30}}    &\underline{\mygreen{47.86}}   \\
    &\%$\Delta\uparrow$ &\cellcolor{gray!20}{\mygreen{15.77}}  &\cellcolor{gray!20}{\mygreen{13.17}} &\cellcolor{gray!20}{\mygreen{14.47}}   &\cellcolor{gray!20}{\mygreen{16.85}}  &\cellcolor{gray!20}{\mygreen{15.55}}  &\cellcolor{gray!20}{\mygreen{16.20}}   &\cellcolor{gray!20}{\mygreen{11.79}} &\cellcolor{gray!20}{\mygreen{10.21}}  &\cellcolor{gray!20}{\mygreen{11.00}}  \\
    \hline
    
    \multirow{2}{*}{ToolRetriever} & ToolBench  &\bf 84.16  &\bf 86.82  &\bf 85.49   &\bf 75.37  &\bf 80.29  &\bf 77.83   &\bf 61.77  &\bf 71.28  &\bf 66.53     \\
    & \texttt{VGToolBench}  &\myred{65.53}   &\myred{70.13}  &\myred{67.83}   &\myred{51.12}  &\myred{56.80}  &\myred{53.96}   &\myred{48.50}  &\myred{56.34}  &\myred{52.42}   \\
    &\%$\Delta\downarrow$   &\cellcolor{gray!20}{\myred{-22.14}}  &\cellcolor{gray!20}{\myred{-19.22}} &\cellcolor{gray!20}{\myred{-20.68}}   &\cellcolor{gray!20}{\myred{-32.17}}  &\cellcolor{gray!20}{\myred{-29.26}}  &\cellcolor{gray!20}{\myred{-30.72}}   &\cellcolor{gray!20}{\myred{-21.48}} &\cellcolor{gray!20}{\myred{-20.96}}  &\cellcolor{gray!20}{\myred{-21.22}}    \\
    \hdashline
    \textbf{\texttt{+\texttt{TRB}}}  & \texttt{VGToolBench} &\underline{\mygreen{75.91}}  &\underline{\mygreen{79.18}}    &\underline{\mygreen{77.55}}   &\underline{\mygreen{63.10}}    &\underline{\mygreen{67.36}}    &\underline{\mygreen{65.23}}   &\underline{\mygreen{57.70}}    &\underline{\mygreen{66.58}}    &\underline{\mygreen{62.14}}   \\
    &\%$\Delta\uparrow$ &\cellcolor{gray!20}{\mygreen{15.84}}  &\cellcolor{gray!20}{\mygreen{12.90}} &\cellcolor{gray!20}{\mygreen{14.37}}   &\cellcolor{gray!20}{\mygreen{23.44}}  &\cellcolor{gray!20}{\mygreen{18.59}}  &\cellcolor{gray!20}{\mygreen{21.01}}   &\cellcolor{gray!20}{\mygreen{18.97}} &\cellcolor{gray!20}{\mygreen{18.18}}  &\cellcolor{gray!20}{\mygreen{18.57}} \\
    \bottomrule
    \end{tabular}}
    \label{tab:main_exp_new}
\end{table*}

\paragraph{\textbf{Baselines}} We use four representative tool retrieval methods as the baseline retrievers in our main experiments, including \textbf{BM25}~\citep{robertson2009probabilistic}, \textbf{TF-IDF}~\citep{ramos2003using}, \textbf{AdaEmbedding} and \textbf{ToolRetriever}~\citep{qin2023toolllm}. Specifically, BM25 and TF-IDF are two classical sparse retrieval methods that estimate the relevance between user instructions and tool documentation based on term frequency statistics. AdaEmbedding and ToolRetriever are two dense retrieval methods that first convert the instructions and tool documentations into semantic embeddings and then calculate the cosine similarity between the resulting embeddings. More specifically, AdaEmbedding uses the OpenAI's \texttt{text-embedding-ada-002} as the embedding model\footnote{\url{https://platform.openai.com/docs/guides/embeddings/embedding-models}}, while ToolRetriever uses the BERT-based embedding model~\citep{kenton2019bert} fine-tuned on the ToolBench dataset.


\paragraph{\textbf{Implementation details}} 

\myred{In our \texttt{TRB}, we employ LLaMA-3.2-3B~\citep{grattafiori2024llama3}, an auto-regressive language model built on an optimized Transformer architecture, as the backbone. We then post-train this backbone using the two-stage training scheme proposed in this work.}
In Stage SFT, the model is supervised fine-tuned using the training dataset of \texttt{VGToolBench} for 3 epochs. In Stage RL, the SFT model is further trained by the DPO algorithm for 3 epochs using the pairwise instruction data.
More specifically, in the pairwise data construction process, we set the number of sampled candidates $N$ to 4. The detailed training hyperparameters for each stage are summarized in Table~\ref{tab:hyperparameter}. All experiments are conducted on two NVIDIA A800 GPUs (80GB). \myred{For the implementation of the training process, we used the public toolkit \texttt{LLama-Factory}\footnote{\url{https://github.com/hiyouga/LLaMA-Factory}} as our code framework.}

\subsection{\textbf{Main Results}}
The main results of tool retrieval are presented in Table~\ref{tab:main_exp_new}. From these results, we can find that:

\textbf{\ding{182} \texttt{TRB} is beneficial to various baseline retrieval methods.} As seen, \texttt{TRB} consistently enhances performance across all baseline retrieval methods. In particular, the improvement with sparse retrieval is significant. Compared to baseline, we see a relative average increase of 111.51\% on \textbf{I2} when using \texttt{TRB} with BM25, highlighting the significant effectiveness of our approach. For dense retrievers, \texttt{TRB} not only helps the ToolRetriever (\textit{i.e.}, BERT-based retriever trained on ToolBench), but is also beneficial to the AdaEmbedding that is based on black-box API. 
Overall, these results suggest that \texttt{TRB} can enhance the preference alignment in both lexical and semantic retrieval paradigms, proving its effectiveness and universality.

\textbf{\ding{183} \texttt{TRB} brings consistent and significant performance gains among all sub-datasets.} 
It can be found that \texttt{TRB} outperforms the baseline retrievers consistently across various sub-datasets of \texttt{VGToolBench}. 
Specifically, taking the BM25 as an example, our \texttt{TRB} brings 87\%, 111.51\%, 36.47\% relative average performance gains against the vanilla retriever for \textbf{I1}, \textbf{I2} and \textbf{I3} subsets, respectively. These results indicate that our \texttt{TRB} works well for single-tool, intra-category and intra-collection multi-tool instruction scenarios, which demonstrate the robustness and generalization capability of \texttt{TRB} across tasks of varying complexity.

\textbf{\ding{184} \texttt{TRB} effectively mitigates the discrepancy between vague user instructions and retriever preferences.} By comparing the performance of original ToolBench and \texttt{VGToolBench}, we see that \texttt{TRB} framework achieves substantial performance gains in both detailed and vague tool scenarios. This indicates that, regardless of whether tools are present in the training set, the bridge model can generate semantically analogous tool names from vague instructions and transforming input sentences into more specific expressions, thereby achieving better alignment with retriever preferences. Moreover, it is noteworthy that, equiped with our \texttt{TRB}, BM25 can even achieve better retrieval performance on \texttt{VGToolBench} than the original ToolBench. We attribute it to the RL optimization, which can better align the retrieval preference and effectively boost the performance of the bridge model.


\subsection{\textbf{Ablation Study}}
\label{sec:ablation}
\myred{In the main experiment above, we have already compared the performance of our model with and without the \texttt{TRB} module. To provide a more comprehensive analysis, we further conduct additional ablation studies in this section. Specifically, we investigate: 1) the effect of different training stages, and 2) the impact of incorporating different RL algorithms within our \texttt{TRB} training processes.}

\paragraph{\textbf{Effect of SFT and RL Training Stages in \texttt{TRB}}} 
As stated in Section~\ref{sec:method}, we employ a two-stage training scheme to post-train the bridge model. In this part, we investigate the individual contributions of each component to the overall performance. Since the initial bridge model does not have the instruction-following ability and is hard to perform well by directly training with RL optimization, we compare the full \texttt{TRB} approach with the variant ``-w/ SFT-only'', \textit{i.e.}, only performing the SFT training process. The comparative results are shown in Table~\ref{tab:ablation_sft}. As seen, compared to the full \texttt{TRB}, ``-w/ SFT-only'' consistently performs worse among all settings. That is, removing the RL training indeed leads to sub-optimal performance, indicating the effectiveness of two-stage training scheme.

\begin{table}[t]
    \centering
    \caption{\textbf{Impact of SFT and RL training stages of \texttt{TRB}.} Notably, taking the BM25 as an example, ``BM25'' means the vanilla tool retriever that does not incorporate with the bridge model. ``-w/ SFT-only'' denotes that we only use the SFT training stage, and "-w/ Full" refers to the full training scheme.}
    \label{tab:ablation_sft}
    \resizebox{\columnwidth}{!}{
    \begin{tabular}{ccccccc}
    \toprule
    \multirow{2}{*}{Method} & \multicolumn{2}{c}{I1} & \multicolumn{2}{c}{I2} & \multicolumn{2}{c}{I3}\\ 
    \cmidrule(lr){2-7}
      & N@5 & N@10 & N@5 & N@10 & N@5 & N@10 \\ 
    \midrule
     \midrule
    BM25 &12.49&13.50&8.81&9.73&13.94&15.70 \\
    \hdashline
    -w/ SFT-only &18.18&19.98&14.13&15.43&16.14&18.86\\
    -w/ Full &\textbf{23.32}&\textbf{25.29}&\textbf{19.06}&\textbf{20.11}&\textbf{18.54}&\textbf{21.97}\\
    \hline
    TF-IDF &31.50&34.19&23.60&25.54&17.64&22.35\\
    \hdashline
    -w/ SFT-only &38.80&42.62&29.82&32.43&25.01&31.38\\
    -w/ Full &\textbf{43.23}&\textbf{46.44}&\textbf{34.56}&\textbf{37.15}&\textbf{27.79}&\textbf{34.38}\\
    \hline
    Ada Embedding &45.15&49.95&27.84&31.07&40.62&45.64\\
    \hdashline
    -w/ SFT-only &50.39&54.97&31.47&34.80&43.47&48.34\\
    -w/ Full &\textbf{52.27}&\textbf{56.53} &\textbf{32.53}&\textbf{35.90}&\textbf{45.41}&\textbf{50.30}\\
    \hline
     ToolRetriever &65.53&70.13&51.12&56.80&48.50&56.34\\
     \hdashline
    -w/ SFT only &74.66&78.22&60.82&65.46&56.23&65.07\\
    -w/ Full &\textbf{75.91}&\textbf{79.18}&\textbf{63.10}&\textbf{67.36}&\textbf{57.70}&\textbf{66.58}\\  
    \bottomrule
    \end{tabular}}
\end{table}

\begin{table}[t]
    \centering
    \caption{\textbf{Performance of various RL algorithms on \texttt{TRB}.} We report tool retrieval scores with \texttt{TRB} under different RL training strategies. ``-w/o RL'' denotes training without reinforcement learning (only using the SFT), while
    ``-w/ *'' refers to using different RL algorithms to continue training the SFT model.
    }
    \label{tab:ablation_rl}
    \resizebox{0.7\columnwidth}{!}{
    \begin{tabular}{ccccc}
    \toprule
    \multirow{2}{*}{Method} & \multicolumn{2}{c}{BM25} & \multicolumn{2}{c}{ToolRetriever} \\
    \cmidrule(lr){2-3} \cmidrule(lr){4-5}
     & N@5 & N@10 & N@5 & N@10 \\
    \midrule
    \midrule
    -w/o RL & 16.14 & 18.86 & 56.23 & 65.07 \\
    \hdashline
    -w/ PPO & 16.24 & 19.48 & 56.87 & 65.95 \\
    -w/ GRPO & 15.87 & 19.00 & 57.09 & 65.98 \\
    -w/ DPO & \textbf{18.54} & \textbf{21.97} & \textbf{57.70} & \textbf{66.58} \\
    \bottomrule
    \end{tabular}}
\end{table}
  
\paragraph{\textbf{Impact of Different RL Algorithms}} To align the \texttt{TRB} with retriever preference more effectively, we explore different RL algorithms, including DPO~\citep{rafailov2023direct}, PPO~\citep{schulman2017proximal}, and GRPO~\citep{shao2024deepseekmath}. These algorithms represent distinct paradigms: DPO operates offline using pairwise preferences, while PPO and GRPO require online reward feedback. 
For the PPO and GRPO algorithms, we directly use the retrieval performance (\textit{i.e.}, NDCG@5) as the reward signal. More specifically, the group size of GRPO is set to 5. To ensure a fair comparison, we adopt the same training hyperparameters (\textit{e.g.}, epoch, learning rate and batch size) for all RL algorithms. As shown in Table~\ref{tab:ablation_rl}, DPO achieves better performance against the other counterparts, while incurring lower computational cost. Notably, PPO and GRPO struggle to achieve desired performance. We conjecture that the ruled-based reward signals is a little sparse and might hard to reinforce the alignment of bridge model. Nevertheless, we believe that a more carefully-designed reward has the great potential for PPO and GRPO to achieve better performance, which is in our future work. Overall, considering the training budgets, we finally choose use the more efficient DPO algorithm in our \texttt{TRB}.

\begin{table}[t]
    \centering
    \caption{\textbf{Performance comparison between different model scales of bridge model.} We report the results of \texttt{TRB} with BM25 and ToolRetriever under \texttt{VGToolBench} (I3). The efficiency is measured by the time consumption to complete one instruction.}
    \resizebox{0.9\columnwidth}{!}{
    \begin{tabular}{ccccccccc}
    \toprule
    \multirow{2}{*}{Method} & \multirow{2}{*}{Model} & \multicolumn{2}{c}{I3} & \multirow{2}{*}{Efficiency} \\ 
    \cmidrule(lr){3-4} 
    & & N@5 & N@10 & \\ 
    \midrule
    \midrule
    \thead{\textbf{w/o \texttt{TRB}}} & - & 13.94 & 15.70 & 0.09s \\
    \cdashline{1-5}
    \multirow{3}{*}{\textbf{\thead{\texttt{TRB} \\ w/ BM25}}} 
    & Llama-3.2-1B & 17.66& 21.55 & 0.43s\\
    & Llama-3.2-3B & \textbf{18.54}& 21.97 & 0.67s\\
    & Llama-3.1-8B & 18.33& \textbf{22.37} & 1.54s\\
    \midrule
    
   \thead{\textbf{w/o \texttt{TRB}}} & - & 48.50 & 56.34  & 0.11s \\
    \cdashline{1-5}
    \multirow{3}{*}{\textbf{\thead{\texttt{TRB} \\ w/ ToolRetriever}}} 
    & Llama-3.2-1B &52.95 & 61.43& 0.52s\\
    & Llama-3.2-3B & 57.70& 66.58 & 0.84s\\
    & Llama-3.1-8B & \textbf{58.94}& \textbf{67.57}&  1.79s\\
    \bottomrule
    \end{tabular}}    
    \label{tab:base_model}
\end{table}

\begin{figure}[t]
    \centering
    \includegraphics[width=0.9\columnwidth]{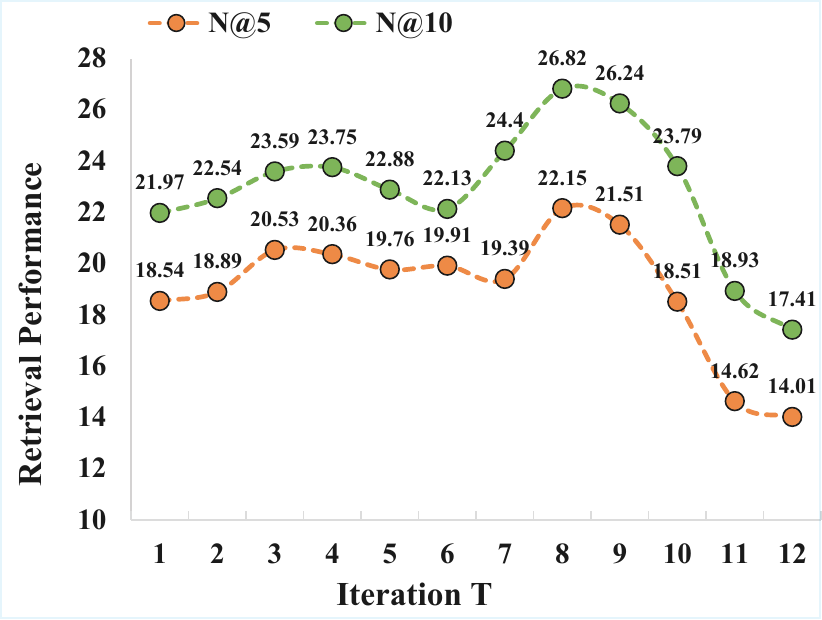}
    
    
    \label{fig:iteration}
    \caption{\myred{\textbf{Effect of the iteration number $T$ in iterative DPO.} We show the retrieval performance (NDCG@5 and NDCG@10) of \texttt{TRB} with BM25 on \texttt{VGToolBench} (\textbf{I3}) across different iterations. Both metrics display a noticeable upward trend followed by a subsequent downturn as $T$ increases, indicating that moderate iterative refinement improves retrieval quality whereas excessive iterations may hinder performance.}}
    
\end{figure}

\subsection{\textbf{In-depth Analysis}}
\label{sec:deep analysis}

\paragraph{\textbf{Influence of Bridge Model Size}} 


\myred{
To assess how the capacity of the bridge model affects the retrieval-bridging mechanism, we conduct experiments across multiple model sizes. We build \texttt{TRB} variants using bridge models from 1B to 8B parameters and evaluate them under both BM25 and ToolRetriever settings. As shown in Table~\ref{tab:base_model}, all \texttt{TRB} variants yield substantial gains over the corresponding ``w/o \texttt{TRB}'' baselines, indicating that the proposed bridging strategy is consistently effective.
Moreover, although larger bridge models can lead to higher retrieval accuracy, \textit{e.g.}, the 8B model achieves the best N@10 results for both retrievers, they will lead to more training and inference budgets. Specifically, when scaling the bridge model from 1B to 8B, the inference latency rises from 0.43 second to 1.54 second under BM25 and from 0.52 second to 1.79 second under ToolRetriever.
Thus, considering both the accuracy improvements and the inference latency, we adopt Llama-3.2-3B as our bridge model, as it achieves a better trade-off between performance and efficiency.
}

\paragraph{\textbf{Effect of Iterative DPO Training}}

It has been demonstrated that DPO effectively enhances model training by better aligning with retriever preferences in above experimental results. The improvement gained through DPO training largely stems from the quality of preference data. Therefore, obtaining more powerful models can generate higher quality preference data, thus facilitating continued DPO training. To explore this, we design an iterative DPO training procedure and investigate the impact of RL iterations $T$ using \texttt{TRB} with BM25. As shown in Figure \ref{fig:iteration}, \texttt{TRB} improves retrieval performance during the first eight iterations. However, increasing the iterations beyond this point does not result in further improvement. Instead, the performance starts to stabilize and even decline, which is similar to the results of~\citet{song2024trial}. One of the possible reasons is that iterative training would hinder the diversity of pairwise data, thus leading to the overfitting of bridge model.
Notably, considering that iterative DPO would lead to much training latency, we set the $T=1$ in the main experiments.

\begin{figure}[ht]
    \centering
    \includegraphics[width=0.85\columnwidth]{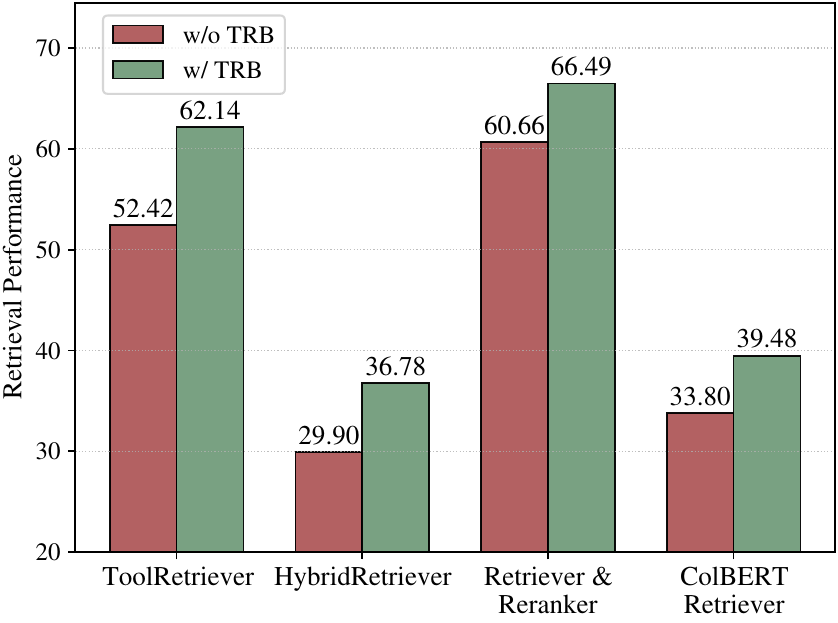}
    \caption{\myred{\textbf{Retrieval performance with state-of-the-art retrieval methods on \texttt{VGToolBench} (I3).} Integrating \texttt{TRB} with the base ToolRetriever surpasses all stronger retrieval methods. Moreover, applying \texttt{TRB} to the hybrid, re-ranking, and ColBERT pipelines yields consistent further improvements.}}
    \label{fig:diff_retriever}
\end{figure}

\begin{figure*}[t]
    \centering
    \includegraphics[width=0.95\textwidth]{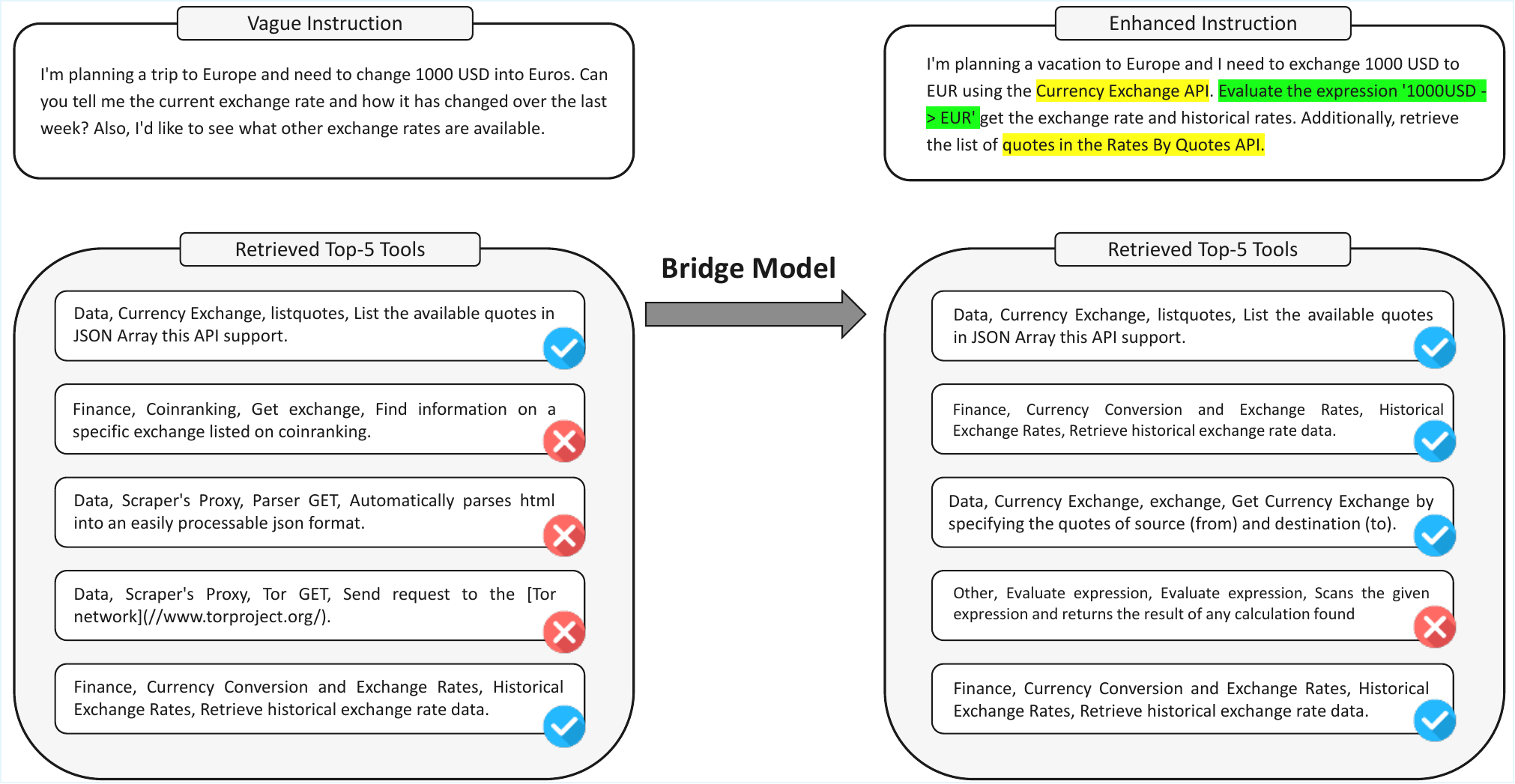}
    \caption{Case study between the vague instruction in \texttt{VGToolBench} and the instruction enhanced via our \texttt{TRB}. In this case, we use the BM25 as the base tool retriever.
    }
    \label{fig:case}
\end{figure*}

\myred{\paragraph{\textbf{Generalization of \texttt{TRB} to State-of-the-Art Retrieval Methods}}
Beyond classical retrievers used in our experiments, more complex paradigms such as hybrid retrieval, re-ranking modules, and ColBERT-style late-interaction architectures are recently incorporated in modern retrieval systems. To rigorously assess the broad applicability of our \texttt{TRB}, we assess it across these cutting-edge retrieval paradigms.
In practice, following the formulation of~\cite{lee2023complementarity}, a \textbf{HybridRetriever} was constructed by linearly combining the scores from the dense ToolRetriever and a sparse BM25 retriever. For the re-ranking pipeline (\textbf{Retriever \& ReRanker}), we employ the powerful \texttt{bge-reranker-v2-m3}\footnote{\url{https://huggingface.co/BAAI/bge-reranker-v2-m3}} to re-order the top-20 candidates retrieved by ToolRetriever and subsequently select the updated top-10. We additionally incorporate \textbf{ColBERTRetriever}, instantiated with ColBERT-v2~\citep{santhanam2022colbertv2}, to represent late-interaction retrieval.
As illustrated in Figure~\ref{fig:diff_retriever}, with the help of our \texttt{TRB}, the base ToolRetriever can achieve much better performance and outperform the other individual retrievers. Moreover, we also apply our \texttt{TRB} to enhance these cutting-edge retrievers, and the results show that \texttt{TRB} is consistently beneficial to all retrievers. For example, for \textbf{HybridRetriever}, \texttt{TRB} brings +6.88\% performance gains. These results demonstrate the effectiveness and generalization of our \texttt{TRB}.
}



\myred{
\paragraph{\textbf{\texttt{TRB} Performance in Real-World Scenarios}}
Since our \texttt{VGToolBench} is a synthetic dataset, some readers may wonder that whether our \texttt{TRB} trained on \texttt{VGToolBench} can work well in real-world scenarios. To verify it, we assess the out-of-distribution (OOD) performance of \texttt{TRB} in a real-world benchmark, \textit{i.e.}, the live subset of the Berkeley Function-Calling Leaderboard (BFCL)~\citep{patil2025the}.
This subset is a human-curated dataset with four configurations (\textit{i.e.}, Simple, Multiple, Parallel, and Parallel Multiple), reflecting increasing levels of tool-use complexity. Notably, the BFCL can be also used to evaluate the tool calling performance of models. Thus, in addition to evaluating the tool retrieval performance of \texttt{TRB}, we also adopt the retrieved results to conduct tool calling task and evaluate its accuracy. As shown in Table~\ref{tab:real_word}, we employ ToolRetriever as the base retriever and QWen2.5-7B-Instruct as the tool generator to evaluate both tool retrieval and tool calling performance. The results indicate that \texttt{TRB} achieves consistent improvement across both tasks. On the one hand, for tool retrieval, \texttt{TRB} achieves +4.45\% average performance gain. Specifically, it leads to +6.18\% and +7.94\% accuracy gains for Parallel and Parallel Multiple settings, respectively. On the other hand, for tool calling, \texttt{TRB} achieves an overall improvement of +5.71\%, demonstrating its consistent advantage across different aspects of tool use. Overall, these results further confirm the generalization of \texttt{TRB} and indicate that \texttt{TRB} has a great potential to improve the subsequent tool calling processes.
}

{%
\arrayrulecolor{black}
\begin{table}[ht]
    \centering
    \caption{\textcolor{black}{\textbf{Performance on the real-world BFCL-live benchmark~\citep{patil2025the}}. We evaluate \texttt{TRB} on both tool retrieval and calling tasks. It can be seen that \texttt{TRB} also works well in this real-world benchmark.}}
    \label{tab:real_word}
    \resizebox{0.95\columnwidth}{!}{%
        \color{black}
        \begin{tabular}{ccccccc}
            \toprule
            \multirow{2}{*}{Task} & \multirow{2}{*}{Method} & \multicolumn{4}{c}{Settings} & \multirow{2}{*}{Overall} \\
            \cmidrule(lr){3-6}
            & & Simple & Multiple & Parallel & \thead{Parallel\\ Multiple} & \\
            \midrule\midrule

            \multirow{3}{*}{\textbf{Tool  Retrieval}} 
            & \thead{\textbf{w/o \texttt{TRB}}} 
                & \textbf{46.68} & 63.54 & 29.52 & 53.57 & 48.33 \\
            \cdashline{2-7}
            & \textbf{\thead{w/ \texttt{TRB}}} 
                & 46.33 & \textbf{67.56} & \textbf{35.70} & \textbf{61.51} & \textbf{52.78} \\
            & \textbf{$\Delta$} 
                & -0.35 & +4.02 & +6.18 & +7.94 & +4.45 \\
            \midrule

            \multirow{3}{*}{\textbf{Tool Calling}} 
            & \thead{\textbf{w/o \texttt{TRB}}} 
                & 51.16 & 57.64 & 37.50 & 50.00 & 49.08 \\
            \cdashline{2-7}
            & \textbf{\thead{w/ \texttt{TRB}}} 
                & \textbf{55.43} & \textbf{59.54} & \textbf{50.00} & \textbf{54.17} & \textbf{54.79} \\
            & \textbf{$\Delta$} 
                & +4.27 & +1.90 & +12.50 & +4.17 & +5.71 \\
            \bottomrule
        \end{tabular}%
    }%
\end{table}
\arrayrulecolor{black}
}%

\subsection{\textbf{Cases Study}}
\label{app:case study}

To have a close look, we provide a case study to illustrate the instruction enhancement process of \texttt{TRB} in Figure \ref{fig:case} .
From this case, we notice that the instruction in \texttt{VGToolBench} is indeed fuzzy and lacks explicit tool information, which is more aligned with read-world user instruction. Directly using the vague instruction to retrieve the Top-5 tools would be hard to achieve optimal performance, underscoring the importance for enhancing the vague instruction. Conversely, with the help of our \texttt{TRB}, the instruction can become more specific (\textit{e.g.,} the phrase "1000USD -> EUR"), which helps the retriever identify more suitable tools (\textit{e.g.,} Data, Currency Exchange). A further advantage of this enhancement is that the relevant tool receives a higher placement in the top-$k$ ranking results, thereby boosting overall retrieval performance.

\section{Conclusion}
\label{sec:conclusion}

In this paper, we focus on the problem of vague instructions in tool retrieval, which is critical yet under-explored. Specifically, we first construct a new tool learning benchmark, \texttt{VGToolBench}, to simulate the vague instructions in real-world scenarios. Based on this, we conduct a series of preliminary analyses and reveal that vague instructions usually damage the performance of tool retrievers. To tackle this issue, we further propose a simple-yet-effective \textbf{Tool Retrieval Bridge} (\texttt{TRB}) approach. In short, \texttt{TRB} introduces a bridge model to align vague instructions with retriever preferences. 
\myred{Our comprehensive experiments show that \texttt{TRB} consistently and significantly outperforms vanilla retrievers, \textit{i.e.}, bringing up to 111.51\% relative average gains, underscoring its effectiveness and generalization.}
\myred{
\section*{Limitations and Future Work} 
While our framework shows clear improvement, several limitations remain. 
First, our \texttt{VGToolBench} is synthesized by a third-party LLM, which may introduce synthetic biases and fail to fully reflect real-world ambiguity. 
Future work will explore more representative data sources, including real user queries and human-in-the-loop refinement. 
In addition, we aim to increase the scale and diversity of the \texttt{VGToolBench} benchmark, incorporating multilingual or domain-specific variations for wider applicability.
Second, incorporating a bridge model into the retrieval pipeline would increase computational cost and may limit scalability in large agent systems. We plan to investigate lighter retrieval architectures and caching strategies to reduce latency.
}

\section*{Acknowledgment}
{
This work was financially supported by the State Grid Corporation Headquarters Science and Technology Project: Research on equipment operation and inspection disposal reasoning technology based on knowledge-enhanced generative model and intelligent agent and demonstration application (5700-202458333A-2-1-ZX).
}

\bibliographystyle{elsarticle-harv} 
\bibliography{reference-revision1}




\end{document}